\pdfoutput=1

\documentclass[11pt]{article}
\usepackage{booktabs}
\usepackage[]{acl}
\usepackage{times}
\usepackage{latexsym}
\usepackage{tabularx}
\usepackage{pgfplots}
\usepackage[T1]{fontenc}

\usepackage[utf8]{inputenc}

\usepackage{microtype}

\usepackage{color}
\usepackage{amsmath}
\usepackage{natbib}
\usepackage{graphicx}
\usepackage{multirow}
\usepackage{nccmath}

\newcommand{\Fig}[1]{Fig.~\ref{#1}}

\newcommand{\Tbl}[1]{Tbl.~\ref{#1}}
\newcommand{\Sec}[1]{Sec.~\ref{#1}}

\newcommand{\ours}{Transkimmer}
\newcommand{\model}[2]{$\text{#1}_{\text{#2}}$}

\title{
\ours{}: Transformer Learns to Layer-wise Skim}

\author{
    Yue Guan\textsuperscript{\rm 1,2,$\star$}, 
    Zhengyi Li\textsuperscript{\rm 1,2,$\star$},
    Jingwen Leng\textsuperscript{\rm 1,2,$\star$}, 
    Zhouhan Lin\textsuperscript{\rm 1,$\mathsection$} \and 
    Minyi Guo\textsuperscript{\rm 1,2,$\dagger$} \\
        \textsuperscript{\rm 1}Shanghai Jiao Tong University, \textsuperscript{\rm 2}Shanghai Qizhi Institute \\ \texttt{\textsuperscript{\rm $\star$}\{bonboru,hobbit,leng-jw\}@sjtu.edu.cn,} \\ \texttt{\textsuperscript{\rm $\mathsection$}lin.zhouhan@gmail.com, \textsuperscript{\rm $\dagger$}guo-my@cs.sjtu.edu.cn}}

\begin{document}
\maketitle

\begin{abstract}

Transformer architecture has become the de-facto model for many machine learning tasks from natural language processing and computer vision.
As such, improving its computational efficiency becomes paramount.
One of the major computational inefficiency of Transformer-based models is that they spend the identical amount of computation throughout all layers.
Prior works have proposed to augment the Transformer model with the capability of skimming tokens to improve its computational efficiency.
However, they suffer from not having effectual and end-to-end optimization of the discrete skimming predictor.
To address the above limitations, we propose the Transkimmer architecture, which learns to identify hidden state tokens that are not required by each layer.
The skimmed tokens are then forwarded directly to the final output, thus reducing the computation of the successive layers.
The key idea in Transkimmer is to add a parameterized predictor before each layer that learns to make the skimming decision.
We also propose to adopt reparameterization trick and add skim loss for the end-to-end training of Transkimmer.
\ours{} achieves $10.97\times$ average speedup on GLUE benchmark compared with vanilla \model{BERT}{base} baseline with less than $1\%$ accuracy degradation. 


\end{abstract}

\section{Introduction} \label{sec:intro}

\begin{figure}[h]
    \centering
    \includegraphics[width=\linewidth]{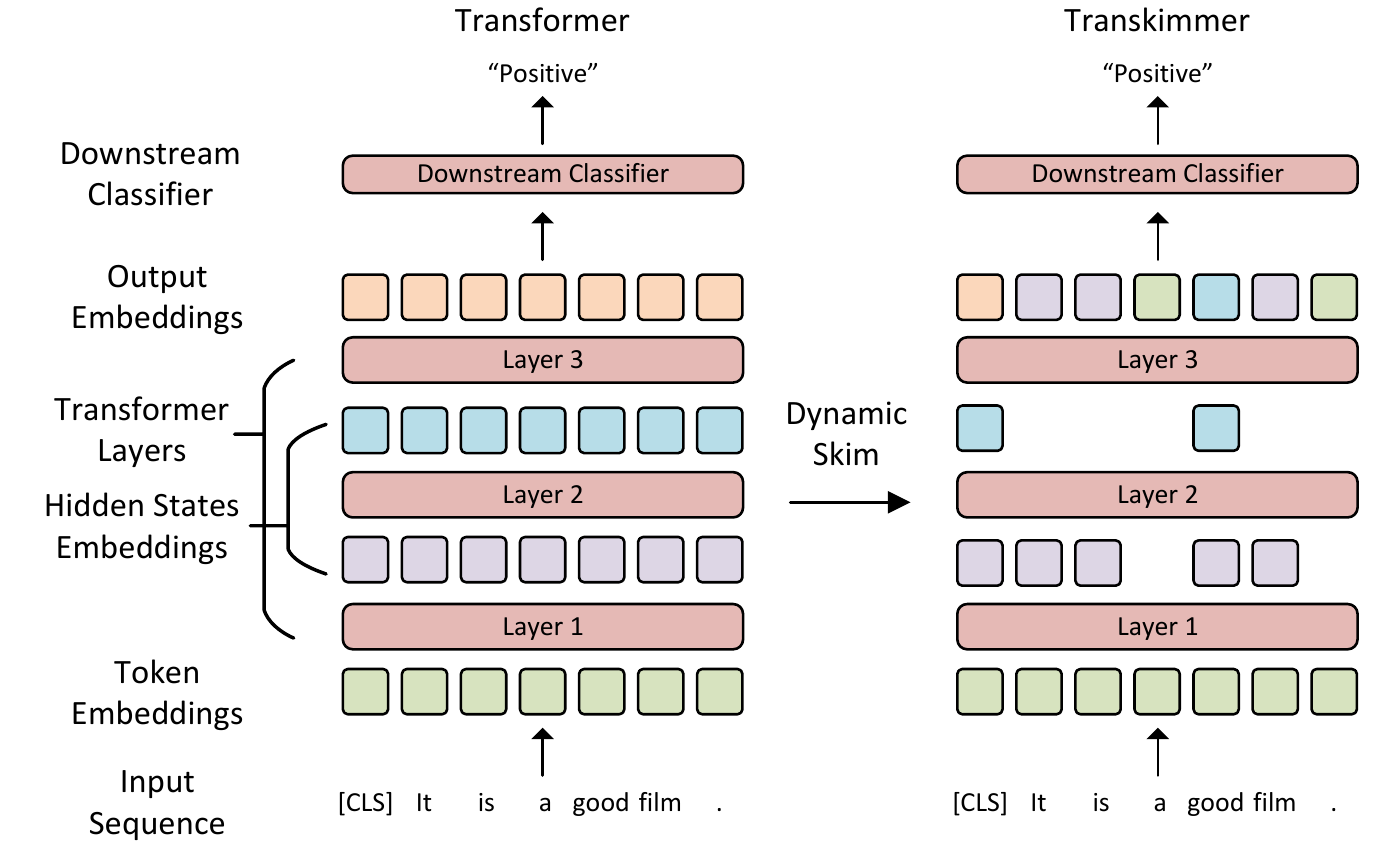}
    \caption{Overview of \ours{} dynamic token skimming method. Tokens are pruned during the processing of Transformer layers. Note that actually we don't need all the tokens given to the downstream classifier in this sequence classification example. We show the full length output embedding sequence  to demonstrate the forwarding design of \ours{}.}
    \label{fig:overview}
\end{figure}

The Transformer model~\cite{vaswani2017attention} has pushed the accuracy of various NLP applications to a new stage by introducing the multi-head attention (MHA) mechanism \cite{lin2017structured}. Further, the BERT~\cite{devlin2018bert} model advances its performances by introducing self-supervised pre-training, and has reached the state-of-the-art accuracy on many NLP tasks. 


Compared to the recurrent fashion models, e.g. RNN~\cite{rumelhart1986learning}, LSTM~\cite{hochreiter1997long}, the Transformer model leverages the above attention mechanism to process all the input sequence.
By doing so, extremely large scale and long span models are enabled, resulting in a huge performance leap in sequence processing tasks.
However, the computation complexity of the attention mechanism is $O(N^2)$ with the input length of $N$, which leads to the high computation demand of the Transformer model. 



Some prior works~\cite{goyal2020power, kim2020length, kim2021learned, ye2021tr} explore the  opportunity on the dynamic reduction of input sequence length to improve the Transformer's computational efficiency.
Its intuition is similar to the human-being's reading comprehension capability that does not read all words equally.
Instead, some words are focused with more interest while others are skimmed.
For Transformer models, this means adopting dynamic computation budget for different input tokens according to their contents.
To excavate the efficiency from this insight, we propose to append a skim predictor module to the Transformer layer to conduct fine-grained dynamic token pruning as shown in \Fig{fig:overview}.
When processed by the Transformer layers, the sequence of token hidden state embeddings are pruned at each layer with reference to its current state.
Less relevant tokens are skimmed without further computation and forwarded to the final output directly.
Only the significant tokens are continued for successive layers for further processing.
This improves the Transformer model inference latency by reducing the input tensors on the sequence length dimension.

However, the optimization problem of such skim decision prediction is non-trivial.
To conduct pruning of dynamic tensors, non-differentiable discrete skim decisions are applied.
Prior works have proposed to use soft-masking approximation or reinforcement learning to resolve, which leads to approximation mismatch or nonuniform optimization.  
\ours{} propose to adopt reparameterization technique~\cite{jang2016categorical} to estimate the gradient for skim prediction.
As such, we can achieve the end-to-end joint optimization obejective and training paradigm.
By jointly training the downstream task and skim objective, the Transformer learns to selectively skim input contents.
In our evaluation, we show \ours{} outperforms all prior input reduction works on inference speedup gain and model accuracy.
Specifically, \model{BERT}{base} is accfelerated for $10.97\times$ on GLUE benchmark and $2.81\times$ without counting the padding tokens.
Moreover, we also demonstrate the method proposed by \ours{} is generally applicable to pre-trained language models and compression methods with RoBERTa, DistillBERT and ALBERT models.

This paper contributes to the following 3 aspects.
\begin{itemize}
\item We propose the \ours{} model which accelerates the Transformer inference with dynamic token skimming.
\item We further propose an end-to-end joint optimization method that trains the skim strategy together with the downstream objective.
\item We evaluate the proposed method on various datasets and backbone models to demonstrate its generality.
\end{itemize}

\section{Related Works}

\paragraph{Recurrent Models with Skimming.}
The idea to skip or skim irrelevant sections or tokens of input sequence has been studied in NLP models, especially recurrent neural networks (RNN)~\cite{rumelhart1986learning} and long short-term memory network (LSTM)~\cite{hochreiter1997long}.
When processed recurrently, skimming the computation of a token is simply jumping the current step and keep the hidden states unchanged.
LSTM-Jump~\cite{yu2017learning}, Skim-RNN~\cite{seo2018neural}, Structural-Jump-LSTM~\cite{hansen2018neural} and  Skip-RNN~\cite{DBLP:journals/corr/abs-1708-06834} adopt this skimming design for acceleration in recurrent models.

\paragraph{Transformer with Input Reduction.}

Unlike the sequential processing of the recurrent models, the Transformer model calculates all the input sequence tokens in parallel.
As such, skimming can be regarded as the reduction of hidden states tensor on sequence length dimension.
Universal Transformer~\cite{dehghani2018universal} proposes a dynamic halting mechanism that determines the refinement steps for each token.
DeFormer~\cite{cao2020deformer} proposes a dual-tower structure to process the question and context part separately at shallow layers specific for QA task.
The context branch is pre-processed off-line and pruned at shallow layers.
Also dedicated for QA tasks, Block-Skim~\cite{guan2021block} proposes to predict and skim the irrelevant context blocks by analyzing the attention weight patterns.
Progressive Growth~\cite{gu2021transformer} randomly drops a portion of input tokens during training to achieve better pre-training efficiency.

Another track of research is to perform such input token selection dynamically during inference, which is the closest to our idea.
POWER-BERT~\cite{goyal2020power} extracts input sequence at token level while processing.
During the fine-tuning process for downstream tasks, \citeauthor{goyal2020power} proposes a soft-extraction layer to train the model jointly.
Length-Adaptive Transformer~\cite{kim2020length} improves it by forwarding the inflected tokens to final downstream classifier as recovery.
Learned Token Pruning~\cite{kim2021learned} improves POWER-BERT by making its pre-defined sparsity ratio a parameterized threshold.
TR-BERT~\cite{ye2021tr} adopts reinforcement learning to independently optimize a policy network that drops tokens.
Comparison to these works are discussed in detail in \Sec{sec:insight}.
Moreover, SpAttn~\cite{wang2021spatten} facilitate POWER-BERT design with a domain-specific hardware design for better acceleration and propose to make skimming decisions with attention values from all layers.

\paragraph{Early Exit}
Early exit~\cite{panda2016conditional, teerapittayanon2016branchynet} is another method to execute the neural network with input-dependent computational complexity.
The idea is to halt the execution during model processing at some early exits.
Under the circumstance of processing sequential inputs, early exit can be viewed as a coarse-grained case of input skimming.
With the hard constraint that all input tokens are skimmed at the same time, early exit methods lead to worse accuracy and performance results compared to input skimming methods.
However, the early exit method is also generally applicable to other domains like convolutional neural networks (CNN).
DeeBERT~\cite{xin2020deebert}, PABEE~\cite{zhou2020bert}, FastBERT~\cite{liu2020fastbert} are some recent works adopting early exit in Transformer models.
Magic Pyramid~\cite{he2021magic} proposes to combine the early exit and the input skimming ideas together.
Tokens are skimmed with fine-grained granularity following POWER-BERT design and the whole input sequence is halted at some early exits.



\paragraph{Efficient Transformer.}
There are also many efforts for designing efficient Transformers~\cite{zhou2020bert,wu2019lite,tay2020efficient}.
For example, researchers have applied well studied compression methods to Transformers, such as pruning~\cite{guo2020accelerating}, quantization~\cite{wang2020q,guo2022squant}, distillation~\cite{sanh2019distilbert}, and weight sharing.
Other efforts focus on dedicated efficient attention mechanism considering its quadratic complexity of sequence length~\cite{kitaev2019reformer,beltagy2020longformer,zaheer2020big} or efficient feed-forward neural network (FFN) design regarding its dominant complexity in Transformer model~\cite{dong2021efficientbert}.
\ours{} is orthogonal to these techniques on the input dimension reduction.


\begin{table}[t]
    \vspace*{0.1cm}
    \centering
    \resizebox{\linewidth}{!}{
    \begin{tabular}{cllll}
    \toprule
      Models         & Optimization                 & Input     & Discard                    & Strategy      \\ \midrule
    
    POWER-BERT    & \multirow{2}{*}{Soft-Masking}                 &  \multirow{2}{*}{Attention} & \multirow{2}{*}{Discard}               & \multirow{2}{*}{Searched} \\
    \cite{goyal2020power} & & & \\\midrule
    
    LAT & \multirow{2}{*}{Soft-Masking}  & \multirow{2}{*}{Attention} & \multirow{2}{*}{Forward}                & \multirow{2}{*}{Searched} \\
    \cite{kim2020length} & & & \\\midrule
    
    LTP     & \multirow{2}{*}{Soft-Masking}     & \multirow{2}{*}{Attention} & \multirow{2}{*}{Discard}               & \multirow{2}{*}{Learned}        \\
    \cite{kim2021learned} & & & \\\midrule
    
    TR-BERT     & \multirow{2}{*}{RL}                 & \multirow{2}{*}{Embedding} & \multirow{2}{*}{Forward}     & \multirow{2}{*}{Searched} \\
    \cite{ye2021tr} & & & \\\midrule
    
    \ours{}       & Reparameterize            & Embedding & Forward             & Learned        \\ \bottomrule
    \end{tabular}
    }
    \caption{Summary of prior token reduction works and their design choices including POWER-BERT, Length-Adaptive Transformer (LAT), Learned Token Pruning (LTP) and TR-BERT. The design details are discussed in \Sec{sec:insight}.}
    \label{table:prior_works}
\end{table}

\section{Input Skimming Search Space} \label{sec:insight}


In this section, we discuss the challenges of dynamic input skimming idea in details.
Moreover, we compare techniques and design decisions from prior works described in \Tbl{table:prior_works}.

\subsection{Optimization Method} \label{sec:optimization_choice}
The first challenge of input skimming is the optimization with discrete skimming decisions.
In specific, the decision for pruning the hidden state tensors (i.e., reducing their sequence length) is a binary prediction. 
As such, the skim prediction model is non-differentiable and unable to be directly optimized by gradient back propagation.
Prior works handle the discrete binary skimming decision by using a set of complicated training techniques, which we categorize in \Tbl{table:prior_works}.


\paragraph{Soft-Masking.}
Some works~\cite{goyal2020power,kim2020length, kim2021learned} propose to use the soft-masking training trick which uses a continuous value for predicting the skimming prediction.
During the training process, the predicted value is multiplied to the hidden states embedding vectors so that no actual pruning happens.
In the inference phase, this continuous skimming prediction value is binarized by a threshold-based step function.
The threshold value is pre-defined or determined through a hyper-parameter search process.
Obviously, there exists a training-inference paradigm mismatch where the actual skimming \textbf{only} happens at the inference time.
Such a mismatch leads to a significant accuracy degradation.

\paragraph{Reinforcement Learning.}
TR-BERT~\cite{ye2021tr} proposes to use the reinforcement learning (RL) to solve the discrete skimming decision problem.
It uses a separated policy network as the skimming predictor, and the backbone Transformer model is considered as the value network.
At first, the backbone Transformer is fine-tuned separately.
It then updates the skimming policy network by using the RL algorithm.
This multi-step training paradigm is tedious.
And training the backbone Transformer and skimming policy network separately is sub-optimal compared to the joint optimization paradigm.
Moreover, the large search space of such RL objective is difficult to converge especially on small downstream datasets.


\paragraph{Reparameterization.} 
In this work, we propose to use the reparameterization technique to address the discrete skimming decision challenge.
Its core idea is to sample the backward propagation gradient during training, whose details we describe in \Sec{sec:method}.
The advantage of our method is that it enables the  joint optimization of skim predictor and backbone Transformer model and therefore achieves the optimal solution.
For example, we will later demonstrate in \Fig{fig:layerwise} that the different tasks or datasets prefer different layer-wise skimming strategies, which are learned by our method.
We will further explain the results in \Sec{sec:post-hoc}.

\begin{figure*}
    \centering
    \includegraphics[width=0.85\linewidth]{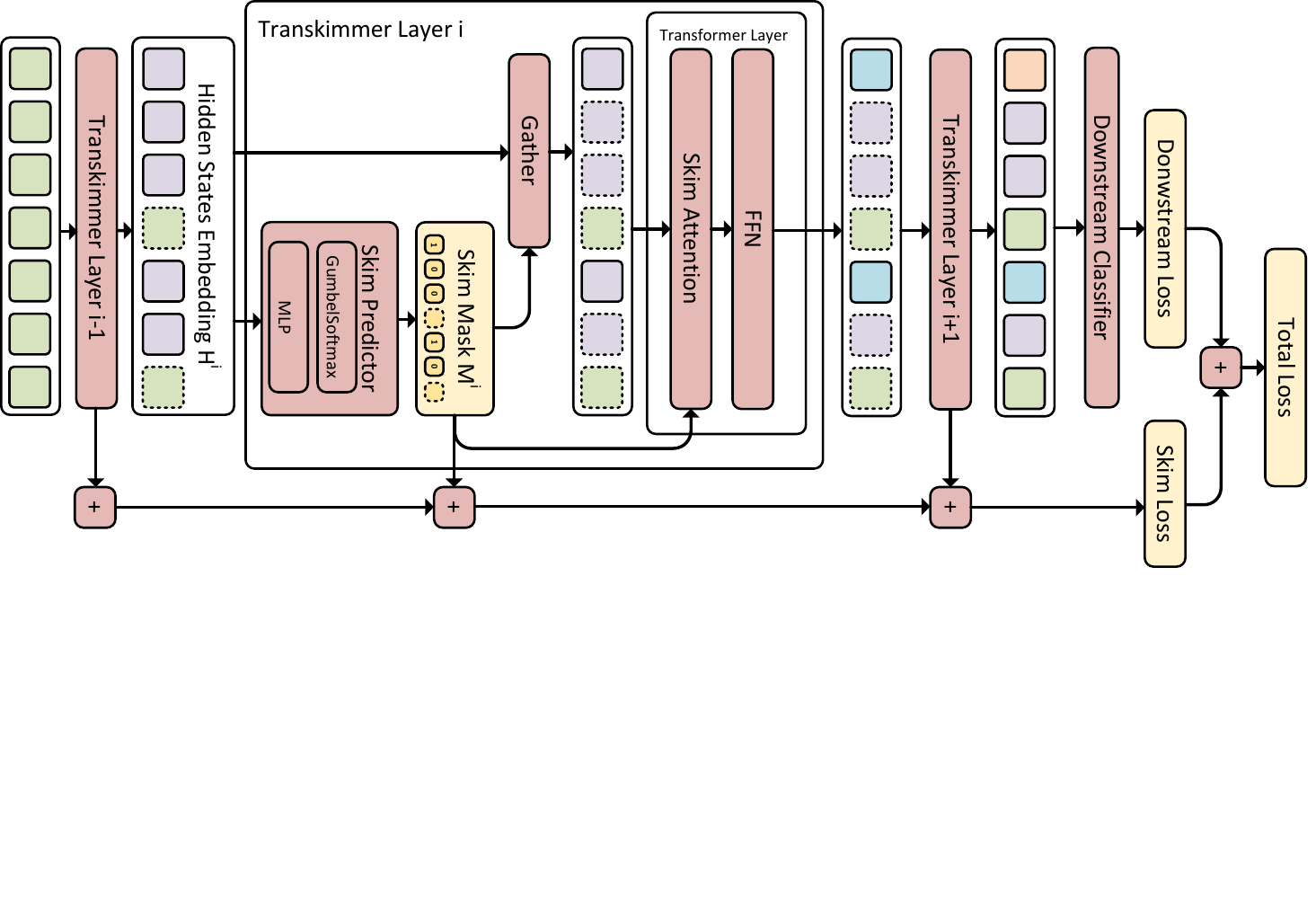}
    \vspace{-3.5cm}
    \caption{Architecture and end-to-end optimization objective of \ours{}. The dashed token embeddings are directly forwarded to the final output of Transformer layers without further processing.}
    \label{fig:arch}
\end{figure*}

\subsection{Design Choices}

In our work, we also jointly consider other design choices regarding the skimming optimization, which includes the choice of input to the skimming module and how to deal with the skimmed input. 
We first explain the choices made by prior works, and then explain the choice of our method.
\paragraph{Strategy.}
For the skimming optimization methods described above, there can be different strategies regarding the implementation details.
Generally, the skimming strategy can be categorized into search-based or learning-based approach, as described in \Tbl{table:prior_works}.
However, when applied to various downstream NLP tasks and datasets, the dynamic skimming scheme prefers different layer-wise strategies as we mentioned above.
This layer-wise skimming characteristics makes the search-based approach not scalable and generally applicable.
In contrast, our method enables the joint training of  skimming strategy and downstream task , which leads to better skimming decisions with reference to both efficiency and accuracy.
LTP is the only by prior works adopting learning-based method, which, however, uses the soft-masking approach and suffers from the training-inference mismatch.



\paragraph{Input for Skimming.}
POWER-BERT, LAT and LTP treat the attention weight value as importance score and utilize it as the criterion for making the skimming decision.
Compared to this value-based method~\cite{guan2020far}, TR-BERT uses hidden state embeddings as input feature.
In our work, we use the hidden state embeddings because they enclose contextual information of the corresponding input token.
Our work shows that the joint training of skimming module and backbone Transformer model leads to that the embeddings also learn to carry features for skimming prediction.

\paragraph{Skimming Tokens.}
For the tokens pruned dynamically by the skimming decision during processing, it is natural to remove them from all the successive layers.
However, LAT and TR-BERT propose to forward such tokens to the final output of the Transformer encoder, which keeps the dimension of the Transformer output unchanged.
Our work adopts the forward-based design because it is more friendly for the Transformer decoder module on downstream tasks.

\section{\ours{} Methodology} \label{sec:method}

\subsection{Transformer with Skim Predictor}

To predict which tokens to be pruned, we append an extra prediction module before each layer as shown in \Fig{fig:arch}.
This prediction module outputs a skimming mask $M$, which is used to gather the hidden state embedding $H$ at the sequence length dimension.
The pruned embedding is then feed to the Transformer layer as its input.
\begin{equation}
\begin{medsize}
\begin{split}
  \scriptsize H^{i+1} &  = \scriptsize Transkimmer^i(H^i) \\
          &  = \scriptsize Transformer^i(Gather(H^i,M^i))  
\end{split}
\end{medsize}
\end{equation}
In the skim mask, we use output 1 to denote remaining tokens and 0 to denote pruned tokens.
The gathering operation is to select the input tensor with a provided mask.
By optimizing this stand-alone skim module, syntactically redundant and semantically irrelevant tokens are skimmed and pruned.
The proposed skim predictor module is a multi-layer perceptron (MLP) network composed of 2 linear layers with a layer norm operation~\cite{ba2016layer} and GeLU activation~\cite{hendrycks2016gaussian}.
The activation function is an arbitrary function with discrete output as skim decision.
\begin{equation} \label{eq:skim_predictor}
\begin{medsize}
\begin{split}
        M^i &= SkimPredictor(H^i) \\
            &= Activation(MLP(H^i)) \\
\text{where } MLP &= Linear(GeLU(LN(Linear))) \\
\end{split}
\end{medsize}
\end{equation}
This skim predictor introduces extra model parameters and computation overhead.
However, both of them are very small compared to the vanilla Transformer model, which are about $7.9\%$ and $6.5\%$ respectively.
We demonstrate later that the computation overhead of skim module is much smaller than the benefits brought by the reduction of input tensor through skimming.

For the tokens pruned by the skim module at each layer, we forward the these pruned hidden state embeddings to the last Transformer layer.
As such, the final output of the whole Transformer model is composed of token embeddings skimmed at all layers and the ones processed by all layers without being skimmed.
\begin{equation}
H^L=\sum_{i=0}^{L-1}H^i\cdot M^i
\end{equation}
And this output is used for classification layers on various downstream tasks.
This makes the skimming operation also compatible for token classification tasks such as extractive question answering (QA) and named entity recognition (NER).
This also restores the once abandoned information for downstream tasks.

\subsection{End-to-End Optimization}

In the above discussion, we have described that \ours{} can be easily augmented to a backbone model without modification to its current structure.
Furthermore, \ours{} is also capable to utilize the pre-trained model parameters and finetune the \ours{} activated Transformer-based models on downstream tasks.
With an extra skim loss appended to the optimization object, this fine-tuning process is also performed end-to-end without changing its origin paradigm.

\paragraph{Skim Attention.}
In the training procedure, \ours{} does not prune the hidden state tensors as it does in the inference time.
Because the gathering and pruning operation of a portion of tokens prevents the back-propagation of their gradients.
The absence of error signal from negative samples interference the convergence of the \ours{} model.
Therefore, we propose skim-attention to mask the reduced tokens in training instead of actually pruning them.
The attention weights to the skimmed tokens are set to 0 and thus unreachable by the other tokens.
\begin{equation}
    SkimAttn(H^i)=Attn(H^i) \cdot M^i
\end{equation}
By doing so, the remaining tokens will have the identical computational value as actually pruning.
And the gradient signal is  passed to the skim predictor module from the skim attention multiplication.

\paragraph{Gumbel Softmax.}
Following the discussion in \Sec{sec:optimization_choice}, the output decision mask of skim predictor is discrete and non-differentiable.
To conquer this inability of back propagation, we use the reparameterization method~\cite{jang2016categorical} to sample the discrete skim prediction from the output probability distribution $\pi^i$ of the MLP.
The gradient of the non-differentiable activation function is estimated from the Gumbel-Softmax distribution during back propagation.
\begin{equation}
\begin{split}
        M^i_j &= Activation(\pi_j^i) \text{     ,for j = 0,1}\\ 
            &= GumbelSoftmax(\pi^i_j)  \\
            &= \frac{exp((log(\pi^i_j)+g^i_j)/\tau)}{\sum_{k=0}^{1}exp((log(\pi^i_k)+g^i_k)/\tau)}
\end{split}
\end{equation}
$g^i_j$ are independent and identically sampled from $Gumbel(0,1)$ distribution. $\tau$ is the temperature hyper-parameter controlling the one-hot prediction distribution. We take $\tau=0.1$ for all experiments.

\begin{table*}[]
    \scriptsize
    \setlength\tabcolsep{2pt}
    \centering
    \resizebox{\linewidth}{!}{
    \begin{tabular}{cccccccccccccc}
    \toprule
    Dataset               & CoLA           & RTE & QQP        & MRPC       & SST-2     & MNLI & WNLI & QNLI & STS-B      & SQuAD & IMDB      & YELP      & 20News \\
    \midrule
    Task                  & Acceptability & NLI & Similarity & Paraphrase & Sentiment & NLI  & NLI & QA   & Similarity & QA    & Sentiment & Sentiment & Sentiment    \\
    Average Sample Length & 11     & 64               & 30    & 53           & 25   & 39          & 37     & 51     & 31           & 152   & 264       & 179       & 551          \\
    Input Sequence Length & 64             & 256 & 128        & 128        & 64        & 128 & 128 & 128  & 64         & 384   & 512       & 512       & 512          \\
    Harmony Coefficient   & 0.3               &  0.8   & 0.2        & 0.5           & 0.3          & 0.2     & 0.5     &  0.1          & 0.3  & 0.8   & 0.5       & 0.5       & 0.5  \\
    \bottomrule
    \end{tabular}
    }
    \caption{Summary of evaluation datasets. The input sequence length matches the setting of prior works POWER-BERT and LTP. It is determined by covering 99 percentile of input samples without truncation.}
    \label{tbl:datasets}
\end{table*}

To achieve better token sparsification ratio, we further add a skim loss term to the overall optimization objective as follows
\begin{equation}
    Loss_{skim} = \frac{1}{L}\sum^{1}_{L-1}\frac{sum(M^i)}{len(M^i)}.
\end{equation}
The skim loss is essentially the ratio of tokens remained in each layer thus representing the computation complexity speedup.
By decreasing this objective, more tokens are forced to be pruned during processing.
To collaborate with the original downstream task loss, we use a harmony coefficient $\lambda$ to balance the two loss terms.
As such, the total loss used for training is formulated as
\begin{equation}
    Loss_{total} = Loss_{downstream} + \lambda Loss_{skim}.
\end{equation}
With the use of the previous settings, the \ours{} model is trained end-to-end without any change to its original training paradigm.

\paragraph{Unbalanced Initialization.}
Another obstacle is that skimming tokens during the training process makes it much unstable and decreases its accuracy performance.
With the pre-trained language modeling parameters, the skim predictor module is random initialized and predicts random decisions.
This induces significant processing mismatch in the backbone Transformer model, where all tokens are accessible.
Consequently, the randomly initialized skim predictor makes the training unstable and diverged.
We propose an unbalance initialization technique to solve this issue.
The idea is to force positive prediction at first and learn to skim gradually.
Generally, parameters are initialized by zero mean distribution as
\begin{equation}
    \omega \sim N(0, \sigma).
\end{equation}
We propose to initialize the bias vector of the last linear layer in the skim predictor MLP with unbalanced bias as
\begin{equation}
    \beta_i \sim N((-1)^{i+1}\mu_0 , \sigma),
\end{equation}
where $i$ stands for the bias vector for prediction 1 or 0.
Consequently, the skim predictor tends to reserve tokens rather than skimming them when innocent.
The mean value $\mu_0$ of the unbalanced distribution set to $5$ for all the experiments.

\section{Evaluation}

\begin{table*}[]
    \scriptsize
        \setlength\tabcolsep{2pt}
      \centering
    \resizebox{\linewidth}{!}{
    \begin{tabular}{lllcccccccccccccccccc}
    \toprule
            & Method           & Padding    & \multicolumn{2}{c}{COLA}      & \multicolumn{2}{c}{RTE}      & \multicolumn{2}{c}{QQP}         & \multicolumn{2}{c}{MRPC}       & \multicolumn{2}{c}{SST-2}       & \multicolumn{2}{c}{MNLI}       & \multicolumn{2}{c}{WNLI}       & \multicolumn{2}{c}{QNLI}       & \multicolumn{2}{c}{STS-B}       \\
            &                  &             & Matthews & FLOPs & Acc. & FLOPs & Acc.  & FLOPs  & F1 & FLOPs & Acc.  & FLOPs & Acc.  & FLOPs & Acc. & FLOPs & Acc. & FLOPs & Pearson & FLOPs \\ \midrule
    \model{BERT}{base}    & Baseline         &    -         & 57.8     & 1.00$\times$  & 65.7 & 1.00$\times$  & 91.3  & 1.00$\times$   & 88.9 & 1.0$\times$  & 93.0    & 1.00$\times$  & 84.9  & 1.00$\times$  & 56.3 & 1.00$\times$  & 91.4 & 1.00$\times$  & 88.6    & 1.00$\times$  \\
            & DeeBERT             & -    & -        &  -    & 66.7    & $1.50\times$     & -     & -      & 85.2    & 1.79$\times$     & 91.5  & 1.89$\times$  & 80.0  & 1.59$\times$  & -    & -     & 87.9    & 1.79$\times$     & -       & -     \\
            & POWER-BERT & Sequence & 52.3     & 4.50$\times$  & 67.4 & 3.40$\times$  & 90.2 & 4.50$\times$   & 88.1 & 2.70$\times$  & 92.1  & 2.40$\times$  & 83.8  & 2.60$\times$  & -    & -     & 90.1 & 2.00$\times$  & 85.1    & 2.00$\times$  \\
            & LAT              & Sequence    & -        & -     & -    & -     & -     & -      & -    & -     & \textbf{92.8}  & 2.90$\times$  & \textbf{84.4}  & 2.80$\times$  & -    & -     & -    & -     & -       & -     \\
            & \ours{}      & No          & 58.9     & 1.75$\times$ & 68.9 & 2.85$\times$ & 90.8  & 2.79$\times$  & 88.5 & 3.13$\times$ & 92.3  & 1.58$\times$ & 83.2  & 2.02$\times$ & 56.3 & 5.56$\times$ & 90.5 & 2.33$\times$ & 87.4    & 3.45$\times$ \\
            & \ours{}      & Sequence    & \textbf{58.9}     & \textbf{18.9$\times$} & \textbf{68.9} & \textbf{4.67$\times$} & \textbf{90.8}  & \textbf{11.72$\times$} & \textbf{88.5} & \textbf{7.45$\times$} & 92.3  & \textbf{10.89$\times$} & 83.2  & \textbf{6.65$\times$} & \textbf{56.3} & \textbf{18.10$\times$} & \textbf{90.5} & \textbf{6.01$\times$} & \textbf{87.4}    & \textbf{18.20$\times$} \\ \cmidrule{2-21} 
            & DistilBERT       &    -         & 55.7     & 1.98$\times$ & 58.8 & 1.98$\times$ & 90.3  & 1.98$\times$  & 88.3 & 1.98$\times$ & 90.6  & 1.98$\times$ & 87.5 & 1.98$\times$ & 53.5 & 1.98$\times$ & 89.3 & 1.98$\times$ & 87.0      & 1.98$\times$ \\
            & \thickspace\thickspace+Transkimmer     & No          & 55.1     & 3.52$\times$ & 59.2 & 4.12$\times$ & 90.1  & 4.95$\times$  & 87.8 & 9.92$\times$ & 89.5  & 5.01$\times$  & 86.7  & 4.40$\times$ & 56.3 & 10.41$\times$ & 87.5 & 4.04$\times$  & 86.5    & 3.47$\times$  \\
            & ALBERT           &   -          & 58.3     & 0.99$\times$ & 70.7 & 0.99$\times$ & 90.2  & 0.99$\times$  & 90.4 & 0.99$\times$ & 90.9  & 0.99$\times$ & 81.8  & 0.99$\times$ & 56.3 & 0.99$\times$ & 89.2 & 0.99$\times$ & 90.4    & 0.99$\times$ \\
            & \thickspace\thickspace+Transkimmer     & No          & 53.4     & 1.52$\times$  & 71.5 & 1.57$\times$  & 90.2  & 3.09$\times$   & 90.6 & 1.94$\times$  & 90.1  & 3.25$\times$  & 81.5  & 1.67$\times$  & 57.7 & 6.19$\times$  & 90.1 & 2.30$\times$   & 89.8    & 1.46$\times$  \\ \midrule
    \model{RoBERTa}{base} & Baseline         &   -          & 61.8     & 1.00$\times$  & 78.0   & 1.00$\times$  & 90.4  & 1.00$\times$   & 92.1 & 1.00$\times$  & 94.3  & 1.00$\times$  & 87.5  & 1.00$\times$  & 56.6 & 1.00$\times$  & 92.9 & 1.00$\times$  & 90.9    & 1.00$\times$  \\
            & LTP              & Batch       & -        & -     & \textbf{78.0}   & 1.81$\times$ & 89.7  & 2.10$\times$  & 91.6 & 2.10$\times$ & \textbf{93.5}  & 2.09$\times$ & 86.5  & 1.88$\times$ & -    & -     & \textbf{92.0}   & 1.87$\times$ & 90.0      & 1.95$\times$ \\
            & \ours{}      & No          & \textbf{61.3}     & \textbf{1.52$\times$} & 76.2 & \textbf{1.79$\times$} & \textbf{91.0}    & \textbf{4.92$\times$}  & \textbf{91.9} & \textbf{2.67$\times$} & \textbf{93.5}  & \textbf{2.08$\times$} & \textbf{86.7}  & \textbf{2.19$\times$} & \textbf{56.3} & \textbf{8.41$\times$}  & 91.7 & \textbf{2.85$\times$} & \textbf{90.5}    & \textbf{2.70$\times$}  \\ \bottomrule
    \end{tabular}
    }
    \caption{Performance and FLOPs (speedup) on GLUE benchmark with \model{BERT}{base} and \model{RoBERTa}{base} as backbone model. \ours{} is adopted on DistilBERT and ALBERT to shows its applicability to general model compression methods.}
    \label{tbl:glue_results}%
    \end{table*}

    \begin{table}[]
    \scriptsize
    \setlength\tabcolsep{2pt}
    \centering
    \resizebox{\linewidth}{!}{
    \begin{tabular}{llcccccccc}
    \toprule
                       &                         & \multicolumn{2}{c}{SQuADv2.0} & \multicolumn{2}{c}{20News} & \multicolumn{2}{c}{Yelp}                     & \multicolumn{2}{c}{IMDB} \\ \cmidrule{3-10} 
    \multirow{-2}{*}{Model} & \multirow{-2}{*}{Padding} & F1             & FLOPs             & Acc.          &  FLOPs          & Acc.                            &       FLOPs     & Acc.         &   FLOPs         \\ \midrule
    \model{BERT}{base}           &                         & 77.1          & 1.00$\times$            & 86.7        & 1.00$\times$           & 69.9                          & 1.00$\times$           & 94.0       & 1.00$\times$          \\
    TR-BERT            & No                      & 75.7          & 2.08$\times$         & \textbf{87.4}        & 4.22$\times$        & 70.0                          & 2.19$\times$        & 93.6        & 2.26$\times$       \\ 
    POWER-BERT         & Sequence                     & -              & -            & 86.5        & 2.91$\times$        & 67.4                          & 2.75$\times$        & 92.1       & 3.05$\times$       \\
    LAT                & Batch                   & -              & -            & -            & -           & -                              & -           & 92.5        & 2.70$\times$        \\ 
    DeFormer           & Sequence                     & 71.4          & 2.19$\times$         & -            & -           & -                              & -           & -           & -          \\ 
    \ours{}         & No                      & \textbf{75.7}          & \textbf{2.10$\times$}          & 86.1        & \textbf{5.27$\times$}        & \textbf{70.1} & \textbf{2.51$\times$} & \textbf{93.7}        & \textbf{2.70$\times$}       \\ \bottomrule
    \end{tabular}
    }
    \caption{Performance and FLOPs evaluation on several downstream tasks and datasets with \model{BERT}{base} as backbone model. The speedup results are emphasized considering the padding setting.}
    \label{tbl:squad_results}
    \end{table}

\subsection{Setup}

\paragraph{Datasets.}

We evaluate the proposed \ours{} method on various datasets.
We use the GLUE\cite{wang2018glue} benchmark including 9 classification/regression datasets, extractive question answering dataset SQuAD-v2.0, and sequence classification datasets 20News~\cite{lang1995newsweeder}, YELP~\cite{zhang2015character} and IMDB~\cite{maas2011learning}.
These datasets are all publicly accessible and the summary is shown in \Tbl{tbl:datasets}. 
The diversity of tasks and text contexts demonstrates the general applicability of the proposed method.

\paragraph{Models.} 
We follow the setting of the BERT model to use the structure of the Transformer encoder and a linear classification layer for all the datasets.
We evaluate the base setting with 12 heads and 12 layers in prior work~\citep{devlin2018bert}.
We implement \ours{} upon BERT and RoBERTa pre-trained language model on downstream tasks.

\paragraph{Baselines.} 
We compare our work to prior token reduction works including POWER-BERT~\cite{goyal2020power}, Length-Adaptive Transformer (LA-Transformer)~\cite{kim2020length}, Learned Token Pruning (LTP)~\cite{kim2021learned}, DeFormer~\cite{cao2020deformer} and TR-BERT~\cite{kim2021learned}.
We also compare our method with model compression methods of knowledge distillation and weight sharing.
Knowledge distillation uses a teacher model to transfer the knowledge to a smaller student model. 
Here we adopt DistilBERT~\citep{sanh2019distilbert} setting to distill a 6-layer model from the \model{BERT}{base} model.
By sharing weight parameters among layers, the amount of weight parameters reduces. 
Note that weight sharing does not impact the computation FLOPs (floating-point operations).
We evaluate \ours{} on ALBERT~\citep{lan2019albert} that shares weight parameters among all layers. 
To express that token reduction method is compatible with these model compression methods, we further implement \ours{} method with this works to demonstrate their cooperation effect.
Besides, DeeBERT\cite{xin2020deebert} is a Transformer early exit baseline which can be regarded as coarse-grained input skimming.

\paragraph{Padding.}
While processing batched input samples, Transformer models perform a padding operation on the input sequences to align the input length.
Sequences are appended with a special padding token \textit{[PAD]} to a predefined sequence length for the convenience of successive computing.
This is a trivial setting for general evaluation but could lead to possible pseudo speedup for token reductions works.
Because the padded tokens can be pruned without prediction.
For the prior works, there are three evaluation settings with reference to padding, padding to a fixed sequence length, padding to mini-batch maximum length and no padding (denoted as \textbf{Sequence}, \textbf{Batch} and \textbf{No} in \Fig{tbl:glue_results} \& \ref{tbl:squad_results}).
We indicate the padding methods of prior works and evaluate \ours{} with different padding settings for a fair comparison.
The speedup of padding to mini-batch maximum length setting is related to batch size and processing order of input samples.
So it is difficult to make a direct comparison under this setting.
However, it can be estimated with padding to fixed sequence length as upper bound and no padding as lower bound.
The sequence length on different datasets is determined following prior works' settings~\cite{goyal2020power,kim2021learned}.
We measure the inference FLOPs as a general measurement of the model computational complexity on all platforms.
We use the TorchProfile\citep{torchprofile} tool to calculate the FLOPs for each model.

\paragraph{Training Setting.} 
We implement the proposed method based on open-sourced library from \citet{Wolf2019HuggingFacesTS}\footnote{The source code is available at \url{https://github.com/ChandlerGuan/Transkimmer}.}.
For each baseline model, we use the released pre-trained checkpoints~\footnote{We use pre-trained checkpoints from \citet{Wolf2019HuggingFacesTS}.}.
We follow the training setting used by \citet{devlin2018bert} and \citet{liu2019roberta} to perform the fine-tuning on the above datasets.
We perform all the experiments reported with random seed 42.
We use four V100 GPUs for training experiments. 

The harmony coefficient $\lambda$ is determined by hyper-parameter grid search on development set with $20\%$ data random  picked from training set set.
The search space is from $0.1$ to $1$ with a step of $0.1$.

\subsection{Overall Results}

We show the overall results on several datasets and demonstrate our observations.
\Tbl{tbl:glue_results} demonstrates the accuracy and speedup evaluated on GLUE benchmark.
And \Tbl{tbl:squad_results} further demonstrates the results on other datasets with longer input.

\paragraph{Comparison to vanilla model baseline.}
Generally, \ours{} achieves considerably speedup to the vanilla models with a minor accuracy degradation, which is less than $1\%$ for nearly all cases.
The average speedup is $2.81\times$ on GLUE benchmark and over $2\times$ on the other datasets.
This demonstrates the inference efficiency improvement of the \ours{} input reduction method.
We also evaluate \ours{} with RoBERTa model as backbone and reach $3.24\times$ average speedup on GLUE benchmark.
This result further expresses the general applicability of \ours{} with different Transformer-based pre-trained language models.
Among all the datasets we evaluated, \ours{} tends to have better acceleration ratio on the easier ones.
For example, sequence classification tasks like QQP and STS-B are better accelerated than QA or NLI datasets. 
We suggest that the Transformer backbone is able to process the information at shallower layers and skim the redundant part earlier.
This is also demonstrated in the following post-hoc analysis \Sec{sec:post-hoc}.

\begin{figure}
    \centering
    \includegraphics[width=0.9\linewidth]{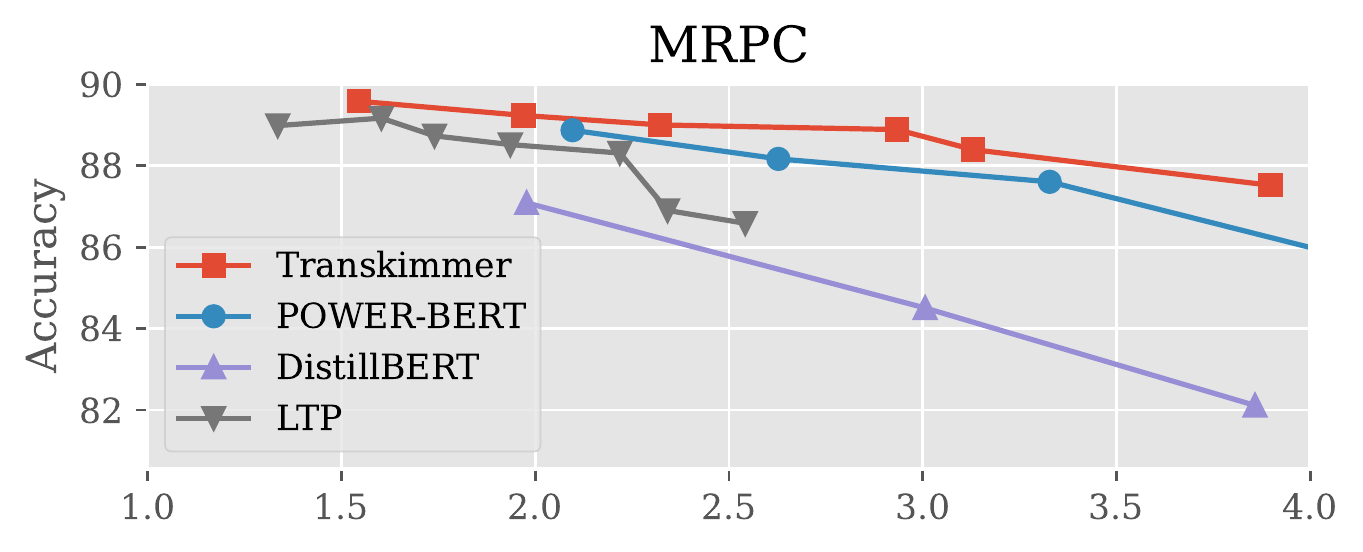}
    \includegraphics[width=0.9\linewidth]{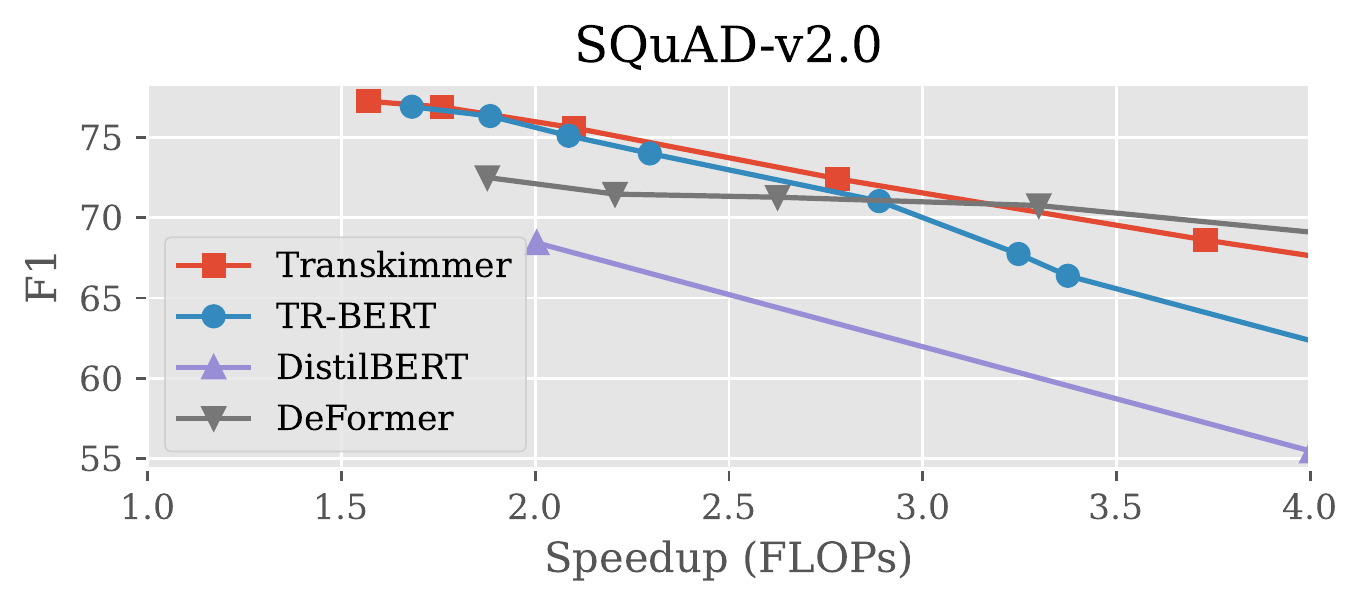}
    \caption{Trade-off results between accuracy and speedup of MRPC and SQuAD-v2.0 datasets by tuning the harmony coefficient. Note that different padding settings are used for each baseline while \ours{} doesn't count any padding. }
    \label{fig:tradeoff}
\end{figure}

\paragraph{Comparison to input reduction prior works.}
As shown in \Tbl{tbl:glue_results}, \ours{} outperforms all the input reduction methods by a margin on GLUE benchmark.
To make a fair comparison, we evaluate \ours{} with two padding settings, padding to fixed sequence length or no padding.
For most cases, \ours{} has better accuracy performance and higher speedup ratio at the same time.
When taking the special padding token into account, \ours{} is able to accelerate \model{BERT}{base} model for $10.97\times$ on GLUE benchmark.
\ours{} also outperforms the other methods on tasks shown in \Tbl{tbl:squad_results}.  
TR-BERT has the closet performance compared with \ours{} but with a much complicated RL paradigm and larger search space.

\paragraph{Comparison to model compression methods.}
The comparison to two model compression methods is shown in \Tbl{tbl:glue_results}.
\ours{} outperforms the knowledge distillation and weight sharing baseline by a margin.
Besides, the dynamic skimming idea itself is orthogonal to this existing model compression methods.
To elaborate, we further adopt the proposed \ours{} method on DistilBERT and ALBERT models.
With the proposed end-to-end training objective, \ours{} is easily augmented to these methods.
There is also no need to change the original training process.
The result shows that the \ours{} method further accelerates the inference efficiency of compressed models with nearly no extra accuracy degradation.

\definecolor{color_0}{RGB}{38,206,205}
\definecolor{color_1}{RGB}{36,185,195}
\definecolor{color_2}{RGB}{35,167,187}
\definecolor{color_3}{RGB}{34,149,179}
\definecolor{color_4}{RGB}{33,133,171}
\definecolor{color_5}{RGB}{31,114,162}
\definecolor{color_6}{RGB}{30,96,154}
\definecolor{color_7}{RGB}{29,82,147}
\definecolor{color_8}{RGB}{28,68,141}
\definecolor{color_9}{RGB}{27,55,135}
\definecolor{color_10}{RGB}{27,55,135}
\definecolor{color_11}{RGB}{27,55,135}
\begin{table*}
  \centering
    \resizebox{\linewidth}{!}{
    \begin{tabular}{cp{17.5cm}c}
    \toprule
    Dataset & Example & \multirow{2}{*}{\begin{tikzpicture}
\begin{axis}[
    hide axis,
    scale only axis,
    height=0pt,
    width=0pt,
    colormap={blackwhite}{ rgb255(0cm)=(38,206,205);rgb255(1cm)=(36,185,195);rgb255(2cm)=(35,167,187);rgb255(3cm)=(34,149,179);rgb255(4cm)=(33,133,171);rgb255(5cm)=(31,114,162);rgb255(6cm)=(30,96,154);rgb255(7cm)=(29,82,147);rgb255(8cm)=(28,68,141);rgb255(9cm)=(27,55,135);rgb255(11cm)=(27,55,135);  rgb255(12cm)=(0,0,0)},
    colorbar ,
    point meta min=1,
    point meta max=12,
    colorbar style={
        width=0.3cm,
        height=5.5cm,
        ytick={1,2,3,4,5,6,7,8,9,10,11,12}
    }]
    \addplot [draw=none] coordinates {(0,0)};
\end{axis}
\end{tikzpicture}}  \\
    \cmidrule{1-2}
    SST-2 & [CLS] \textcolor{color_10}{Even} \textcolor{color_2}{horror} \textcolor{color_2}{fans} \textcolor{color_5}{will} \textcolor{color_3}{most} \textcolor{color_7}{likely} \textcolor{color_11}{not} \textcolor{color_8}{find} \textcolor{color_7}{what} \textcolor{color_2}{they}\textcolor{color_1}{'}\textcolor{color_2}{re} \textcolor{color_5}{seeking} \textcolor{color_1}{with} \textcolor{color_10}{trouble} \textcolor{color_4}{every} \textcolor{color_4}{day}\textcolor{color_7}{;} \textcolor{color_1}{the} \textcolor{color_2}{movie} \textcolor{color_10}{lacks} \textcolor{color_2}{both} \textcolor{color_4}{thrill}\textcolor{color_1}{s} \textcolor{color_1}{and} \textcolor{color_2}{humor}\textcolor{color_2}{.} \textcolor{color_0}{[SEP]} \\
    \cmidrule{1-2}
    \multirow{2}[2]{*}{SQuAD} & Question: [CLS] \textcolor{color_10}{In} what country \textcolor{color_8}{is} Normandy \textcolor{color_10}{located}? \textcolor{color_2}{[SEP]}  \\
       & Context: \textcolor{color_11}{The} Norman\textcolor{color_11}{s} \textcolor{color_6}{(}\textcolor{color_10}{Norman}\textcolor{color_4}{:} \textcolor{color_8}{No}\textcolor{color_4}{ur}\textcolor{color_4}{man}\textcolor{color_6}{ds}\textcolor{color_6}{;} \textcolor{color_7}{French}\textcolor{color_4}{:} \textcolor{color_7}{Norman}\textcolor{color_6}{ds}\textcolor{color_6}{;} \textcolor{color_6}{Latin}\textcolor{color_4}{:} \textcolor{color_8}{Norman}\textcolor{color_6}{ni}\textcolor{color_7}{)} \textcolor{color_4}{were} \textcolor{color_7}{the} \textcolor{color_8}{people} \textcolor{color_4}{who} \textcolor{color_7}{in} \textcolor{color_6}{the} \textcolor{color_10}{10th} \textcolor{color_6}{and} \textcolor{color_6}{11th} \textcolor{color_10}{centuries} \textcolor{color_4}{gave} \textcolor{color_4}{their} \textcolor{color_4}{name} \textcolor{color_6}{to} Normandy\textcolor{color_9}{,} a region in France\textcolor{color_11}{.} \textcolor{color_8}{They} \textcolor{color_3}{were} \textcolor{color_8}{descended} \textcolor{color_4}{from} \textcolor{color_10}{Norse} \textcolor{color_4}{(}\textcolor{color_5}{"}\textcolor{color_8}{Norman}\textcolor{color_6}{"} \textcolor{color_6}{comes} \textcolor{color_4}{from} \textcolor{color_6}{"}\textcolor{color_6}{Norse}\textcolor{color_6}{man}\textcolor{color_7}{"}\textcolor{color_6}{)} \textcolor{color_7}{raiders} \textcolor{color_4}{and} \textcolor{color_11}{pirates} \textcolor{color_4}{from} \textcolor{color_10}{Denmark}\textcolor{color_4}{,} \textcolor{color_8}{Iceland} \textcolor{color_4}{and} \textcolor{color_10}{Norway} \textcolor{color_4}{who}\textcolor{color_4}{,} \textcolor{color_4}{under} \textcolor{color_3}{their} \textcolor{color_4}{leader} \textcolor{color_6}{Roll}\textcolor{color_6}{o}\textcolor{color_4}{,} \textcolor{color_4}{agreed} \textcolor{color_4}{to} \textcolor{color_4}{swear} \textcolor{color_4}{fe}\textcolor{color_4}{al}\textcolor{color_4}{ty} \textcolor{color_4}{to} \textcolor{color_5}{King} \textcolor{color_4}{Charles} \textcolor{color_4}{III} \textcolor{color_3}{of} \textcolor{color_5}{West} \textcolor{color_4}{Fran}\textcolor{color_6}{cia}\textcolor{color_6}{.} \textcolor{color_4}{Through} \textcolor{color_4}{generations} \textcolor{color_1}{of} \textcolor{color_4}{assimilation} \textcolor{color_3}{and} \textcolor{color_4}{mixing} \textcolor{color_3}{with} \textcolor{color_4}{the} \textcolor{color_4}{native} \textcolor{color_5}{Frankish} \textcolor{color_3}{and} \textcolor{color_4}{Roman}\textcolor{color_3}{-}\textcolor{color_4}{Gaul}\textcolor{color_4}{ish} \textcolor{color_6}{populations}\textcolor{color_4}{,} \textcolor{color_4}{their} \textcolor{color_4}{descendants} \textcolor{color_4}{would} \textcolor{color_4}{gradually} \textcolor{color_4}{merge} \textcolor{color_3}{with} \textcolor{color_4}{the} \textcolor{color_4}{Carol}\textcolor{color_4}{ing}\textcolor{color_4}{ian}\textcolor{color_3}{-}\textcolor{color_4}{based} \textcolor{color_4}{cultures} \textcolor{color_2}{of} \textcolor{color_4}{West} \textcolor{color_4}{Fran}\textcolor{color_6}{cia}\textcolor{color_4}{.} \textcolor{color_4}{The} \textcolor{color_4}{distinct} \textcolor{color_4}{cultural} \textcolor{color_3}{and} \textcolor{color_4}{ethnic} \textcolor{color_4}{identity} \textcolor{color_1}{of} \textcolor{color_4}{the} \textcolor{color_6}{norman}\textcolor{color_4}{s} \textcolor{color_4}{emerged} \textcolor{color_4}{initially} \textcolor{color_4}{in} \textcolor{color_4}{the} \textcolor{color_5}{first} \textcolor{color_5}{half} \textcolor{color_1}{of} \textcolor{color_4}{the} \textcolor{color_4}{10th} \textcolor{color_6}{century}\textcolor{color_4}{,} \textcolor{color_4}{and} \textcolor{color_4}{it} \textcolor{color_4}{continued} \textcolor{color_3}{to} \textcolor{color_4}{evolve} \textcolor{color_4}{over} \textcolor{color_4}{the} \textcolor{color_4}{succeeding} \textcolor{color_4}{centuries}\textcolor{color_3}{.} \textcolor{color_2}{[SEP]}  \\
       & Answer: France \\
    \bottomrule
\end{tabular}
}
    \caption{Post-hoc case study of SST-2 sentimental analysis and SQuAD QA tasks from \ours{} model with \model{BERT}{base} setting. The color indicated by the colorbar represents the Transformer layer index where the token is pruned. Specifically, the black tokens are fully processed without being skimmed.}
    \label{tbl:post-hoc}
\end{table*}

\subsection{Accuracy and Performance Trade-Off}
\Fig{fig:tradeoff} demonstrates the accuracy and performance trade-off analysis by tuning the harmony coefficient.
We show the results on MRPC and SQuAD-v2.0 datasets to give comparisons with different baselines.
It is shown that \ours{} achieves a better accuracy to speedup Pareto curve compared to prior works.
\ours{} is able to provide better acceleration gain with less accuracy degradation.
Especially, \ours{} has a $1.5\times$ speedup without accuracy loss.
The result validates our design decisions analyzed in the input reduction search space choices.

\begin{figure}
    \centering
    \includegraphics[width=\linewidth]{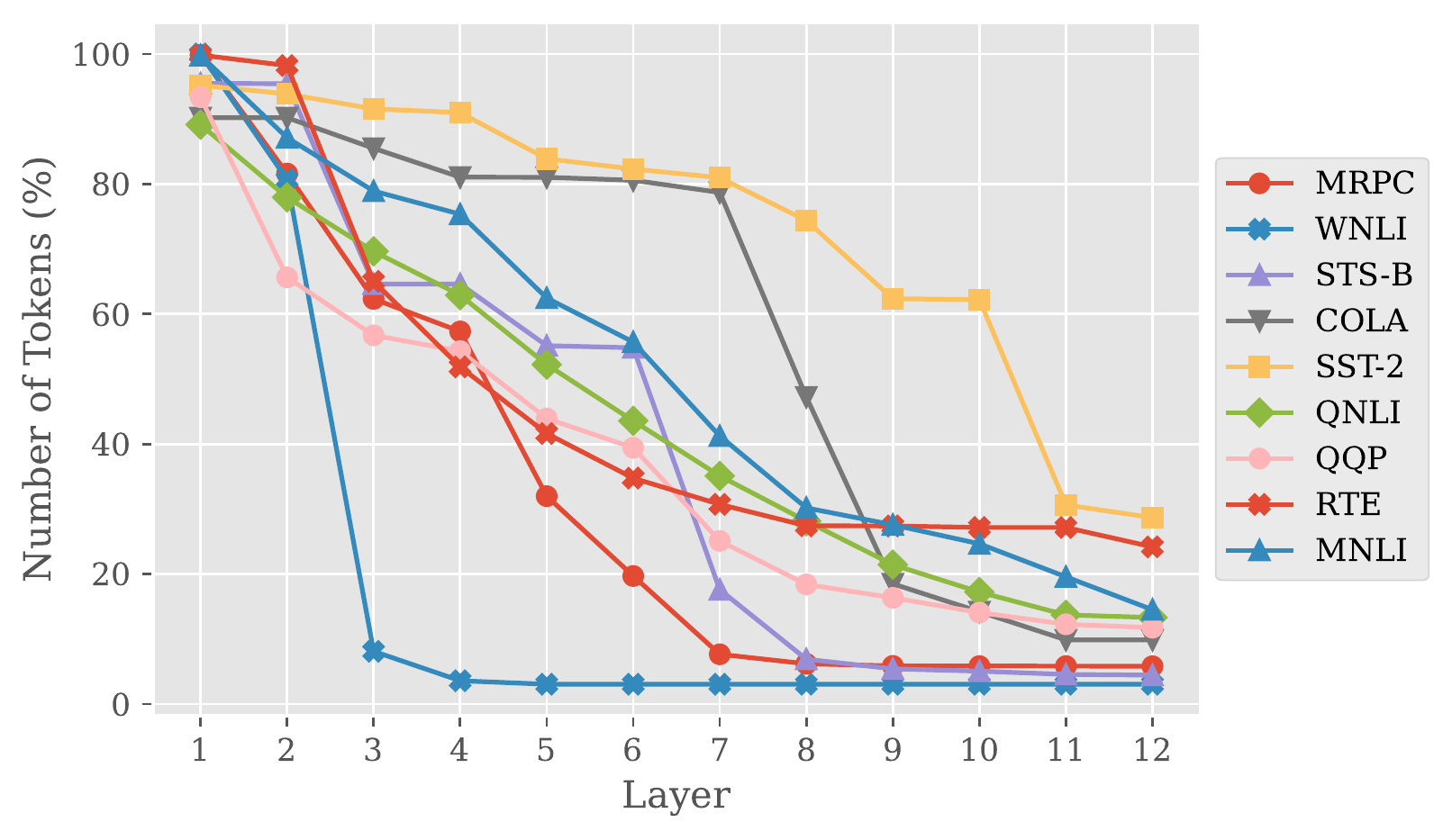}
    \caption{Layer-wise skim strategies analysis of datasets from GLUE benchmark. The normalized area under curve is viewed as an approximate speedup ratio with reference to sequence length.}
    \label{fig:layerwise}
\end{figure}

\subsection{Post-hoc Analysis} \label{sec:post-hoc}
\paragraph{Skim Strategy.} 
\Fig{fig:layerwise} is the result of the number of tokens remained for the processing of each Transformer layer.
The normalized area under each curve is a rough approximation of the speedup ratio with reference to the tokens number.
By end-to-end optimization, \ours{} learns significant distinguished strategies on different tasks.
On WNLI dataset, over $90\%$ of tokens are pruned within the first 3 layers and guarantees a high acceleration gain.
The steer cliff at layer 7 on COLA demonstrates a large portion of skimming at this particular position.
We suggest that this is because the processing of contextual information is sufficient for the skimming decision at this specific layer.

\paragraph{Post-Hoc Case Study.} 
Moreover, several post-hoc case studies are demonstrated with \Tbl{tbl:post-hoc}.
In the SST-2 sentimental analysis example, the definite articles and apostrophes are discarded at the beginning.
And all words are encoded in contextual hidden states embeddings and gradually discarded except for a few significant key words.
Only the special token \textit{[CLS]} is fully processed in this example for final sentimental classification.
However, on the token classification task example from SQuAD dataset, all tokens are given to the downstream classifier to predict the answer position.
The answer tokens are processed by all Transformer layers.
Similarly, the question part is also kept with tokens containing enough  information.
Another detail worth mentioning is that we use subword tokenization for the SQuAD dataset.
As such, subword tokens of the same word might be discarded at different layers.
For instance, the word \textit{Francia} is tokenized into \textit{fran-} and \textit{-cia} two subword tokens, which are pruned at layer 4 and 6 respectively.

\section{Conclusion} \label{sec:conclusion}

Input skimming or dynamic input reduction is an emerging Transformer model acceleration method studied by many works recently. 
This idea utilizes the semantic structure of language and the syntactic information of the input context for inference acceleration.
Compared to static model weight compression methods, input skimming explores the redundancy in the input and hidden state tensors.
As such, it is orthogonal and compatible with those model compression algorithms with its dynamic feature.

In this work, we propose an accurate and efficient Transformer inference acceleration method by teaching it how to skim input contents.
The proposed \ours{} method is trained with an easy and end-to-end paradigm.
Furthermore, \ours{} is also generally applicable to various Transformer-based model structures.
It is even compatible with the static model compression methods like knowledge distillation and weight sharing.
We believe that the above features guarantee the \ours{} method a wide range of applicable production scenarios.


\section*{Acknowledgement}
This work was supported by the National Key R\&D Program of China under Grant 2021ZD0110104, the National Natural Science Foundation of China (NSFC) grant (U21B2017, 62106143, 62072297, and 61832006), and Shanghai Pujiang Program.
We would like to thank the reviewers of ACL rolling review for their supportive comments and suggestions.
Jingwen Leng and Minyi Guo are the corresponding authors of this paper.

\bibliography{bibliography}

\begin{thebibliography}{45}
\expandafter\ifx\csname natexlab\endcsname\relax\def\natexlab#1{#1}\fi

\bibitem[{Ba et~al.(2016)Ba, Kiros, and Hinton}]{ba2016layer}
Jimmy~Lei Ba, Jamie~Ryan Kiros, and Geoffrey~E Hinton. 2016.
\newblock \href {https://arxiv.org/abs/1607.06450} {Layer normalization}.
\newblock \emph{arXiv preprint arXiv:1607.06450}.

\bibitem[{Beltagy et~al.(2020)Beltagy, Peters, and
  Cohan}]{beltagy2020longformer}
Iz~Beltagy, Matthew~E Peters, and Arman Cohan. 2020.
\newblock \href {https://arxiv.org/abs/2004.05150} {Longformer: The
  long-document transformer}.
\newblock \emph{arXiv preprint arXiv:2004.05150}.

\bibitem[{Campos et~al.(2018)Campos, Jou, Gir{\'{o}}{-}i{-}Nieto, Torres, and
  Chang}]{DBLP:journals/corr/abs-1708-06834}
V{\'{\i}}ctor Campos, Brendan Jou, Xavier Gir{\'{o}}{-}i{-}Nieto, Jordi Torres,
  and Shih{-}Fu Chang. 2018.
\newblock \href {https://openreview.net/forum?id=HkwVAXyCW} {Skip {RNN:}
  learning to skip state updates in recurrent neural networks}.
\newblock In \emph{6th International Conference on Learning Representations,
  {ICLR} 2018, Vancouver, BC, Canada, April 30 - May 3, 2018, Conference Track
  Proceedings}. OpenReview.net.

\bibitem[{Cao et~al.(2020)Cao, Trivedi, Balasubramanian, and
  Balasubramanian}]{cao2020deformer}
Qingqing Cao, Harsh Trivedi, Aruna Balasubramanian, and Niranjan
  Balasubramanian. 2020.
\newblock \href {https://doi.org/10.18653/v1/2020.acl-main.411} {{D}e{F}ormer:
  Decomposing pre-trained transformers for faster question answering}.
\newblock In \emph{Proceedings of the 58th Annual Meeting of the Association
  for Computational Linguistics}, pages 4487--4497, Online. Association for
  Computational Linguistics.

\bibitem[{Dehghani et~al.(2019)Dehghani, Gouws, Vinyals, Uszkoreit, and
  Kaiser}]{dehghani2018universal}
Mostafa Dehghani, Stephan Gouws, Oriol Vinyals, Jakob Uszkoreit, and Lukasz
  Kaiser. 2019.
\newblock \href {https://openreview.net/forum?id=HyzdRiR9Y7} {Universal
  transformers}.
\newblock In \emph{7th International Conference on Learning Representations,
  {ICLR} 2019, New Orleans, LA, USA, May 6-9, 2019}. OpenReview.net.

\bibitem[{Devlin et~al.(2019)Devlin, Chang, Lee, and
  Toutanova}]{devlin2018bert}
Jacob Devlin, Ming-Wei Chang, Kenton Lee, and Kristina Toutanova. 2019.
\newblock \href {https://doi.org/10.18653/v1/N19-1423} {{BERT}: Pre-training of
  deep bidirectional transformers for language understanding}.
\newblock In \emph{Proceedings of the 2019 Conference of the North {A}merican
  Chapter of the Association for Computational Linguistics: Human Language
  Technologies, Volume 1 (Long and Short Papers)}, pages 4171--4186,
  Minneapolis, Minnesota. Association for Computational Linguistics.

\bibitem[{Dong et~al.(2021)Dong, Wang, Xu, Peng, Ren, and
  Liang}]{dong2021efficientbert}
Chenhe Dong, Guangrun Wang, Hang Xu, Jiefeng Peng, Xiaozhe Ren, and Xiaodan
  Liang. 2021.
\newblock Efficientbert: Progressively searching multilayer perceptron via
  warm-up knowledge distillation.
\newblock In \emph{Findings of the Association for Computational Linguistics:
  EMNLP 2021}, pages 1424--1437.

\bibitem[{Goyal et~al.(2020)Goyal, Choudhury, Raje, Chakaravarthy, Sabharwal,
  and Verma}]{goyal2020power}
Saurabh Goyal, Anamitra~Roy Choudhury, Saurabh Raje, Venkatesan~T.
  Chakaravarthy, Yogish Sabharwal, and Ashish Verma. 2020.
\newblock \href {http://proceedings.mlr.press/v119/goyal20a.html} {Power-bert:
  Accelerating {BERT} inference via progressive word-vector elimination}.
\newblock In \emph{Proceedings of the 37th International Conference on Machine
  Learning, {ICML} 2020, 13-18 July 2020, Virtual Event}, volume 119 of
  \emph{Proceedings of Machine Learning Research}, pages 3690--3699. {PMLR}.

\bibitem[{Gu et~al.(2021)Gu, Liu, Yu, Li, Chen, and Han}]{gu2021transformer}
Xiaotao Gu, Liyuan Liu, Hongkun Yu, Jing Li, Chen Chen, and Jiawei Han. 2021.
\newblock \href {https://doi.org/10.18653/v1/2021.naacl-main.406} {On the
  transformer growth for progressive {BERT} training}.
\newblock In \emph{Proceedings of the 2021 Conference of the North American
  Chapter of the Association for Computational Linguistics: Human Language
  Technologies}, pages 5174--5180, Online. Association for Computational
  Linguistics.

\bibitem[{Guan et~al.(2020)Guan, Leng, Li, Chen, and Guo}]{guan2020far}
Yue Guan, Jingwen Leng, Chao Li, Quan Chen, and Minyi Guo. 2020.
\newblock \href {https://doi.org/10.18653/v1/2020.coling-main.342} {How far
  does {BERT} look at: Distance-based clustering and analysis of {BERT}{'}s
  attention}.
\newblock In \emph{Proceedings of the 28th International Conference on
  Computational Linguistics}, pages 3853--3860, Barcelona, Spain (Online).
  International Committee on Computational Linguistics.

\bibitem[{Guan et~al.(2021)Guan, Li, Leng, Lin, Guo, and Zhu}]{guan2021block}
Yue Guan, Zhengyi Li, Jingwen Leng, Zhouhan Lin, Minyi Guo, and Yuhao Zhu.
  2021.
\newblock \href {https://arxiv.org/abs/2112.08560} {Block-skim: Efficient
  question answering for transformer}.
\newblock \emph{arXiv preprint arXiv:2112.08560}.

\bibitem[{Guo et~al.()Guo, Hsueh, Leng, Qiu, Guan, Wang, Jia, Li, Guo, and
  Zhu}]{guo2020accelerating}
Cong Guo, Bo~Hsueh, Jingwen Leng, Yuxian Qiu, Yue Guan, Zehuan Wang, Xiaoying
  Jia, Xipeng Li, Minyi Guo, and Yuhao Zhu.
\newblock Accelerating sparse dnn models without hardware-support via tile-wise
  sparsity.
\newblock In \emph{2020 SC20: International Conference for High Performance
  Computing, Networking, Storage and Analysis (SC)}, pages 204--218. IEEE
  Computer Society.

\bibitem[{Guo et~al.(2022)Guo, Qiu, Leng, Gao, Zhang, Liu, Yang, Zhu, and
  Guo}]{guo2022squant}
Cong Guo, Yuxian Qiu, Jingwen Leng, Xiaotian Gao, Chen Zhang, Yunxin Liu, Fan
  Yang, Yuhao Zhu, and Minyi Guo. 2022.
\newblock \href {https://openreview.net/forum?id=JXhROKNZzOc} {{SQ}uant:
  On-the-fly data-free quantization via diagonal hessian approximation}.
\newblock In \emph{International Conference on Learning Representations}.

\bibitem[{Hansen et~al.(2019)Hansen, Hansen, Alstrup, Simonsen, and
  Lioma}]{hansen2018neural}
Christian Hansen, Casper Hansen, Stephen Alstrup, Jakob~Grue Simonsen, and
  Christina Lioma. 2019.
\newblock \href {https://openreview.net/forum?id=B1xf9jAqFQ} {Neural speed
  reading with structural-jump-lstm}.
\newblock In \emph{7th International Conference on Learning Representations,
  {ICLR} 2019, New Orleans, LA, USA, May 6-9, 2019}. OpenReview.net.

\bibitem[{He et~al.(2021)He, Keivanloo, Xu, He, Zeng, Rajagopalan, and
  Chilimbi}]{he2021magic}
Xuanli He, Iman Keivanloo, Yi~Xu, Xiang He, Belinda Zeng, Santosh Rajagopalan,
  and Trishul Chilimbi. 2021.
\newblock \href {https://arxiv.org/abs/2111.00230} {Magic pyramid: Accelerating
  inference with early exiting and token pruning}.
\newblock \emph{arXiv preprint arXiv:2111.00230}.

\bibitem[{Hendrycks and Gimpel(2016)}]{hendrycks2016gaussian}
Dan Hendrycks and Kevin Gimpel. 2016.
\newblock \href {https://arxiv.org/abs/1606.08415} {Gaussian error linear units
  (gelus)}.
\newblock \emph{arXiv preprint arXiv:1606.08415}.

\bibitem[{Hochreiter and Schmidhuber(1997)}]{hochreiter1997long}
Sepp Hochreiter and J{\"u}rgen Schmidhuber. 1997.
\newblock Long short-term memory.
\newblock \emph{Neural computation}, 9(8):1735--1780.

\bibitem[{Jang et~al.(2017)Jang, Gu, and Poole}]{jang2016categorical}
Eric Jang, Shixiang Gu, and Ben Poole. 2017.
\newblock \href {https://openreview.net/forum?id=rkE3y85ee} {Categorical
  reparameterization with gumbel-softmax}.
\newblock In \emph{5th International Conference on Learning Representations,
  {ICLR} 2017, Toulon, France, April 24-26, 2017, Conference Track
  Proceedings}. OpenReview.net.

\bibitem[{Kim and Cho(2021)}]{kim2020length}
Gyuwan Kim and Kyunghyun Cho. 2021.
\newblock \href {https://doi.org/10.18653/v1/2021.acl-long.508}
  {Length-adaptive transformer: Train once with length drop, use anytime with
  search}.
\newblock In \emph{Proceedings of the 59th Annual Meeting of the Association
  for Computational Linguistics and the 11th International Joint Conference on
  Natural Language Processing (Volume 1: Long Papers)}, pages 6501--6511,
  Online. Association for Computational Linguistics.

\bibitem[{Kim et~al.(2021)Kim, Shen, Thorsley, Gholami, Kwon, Hassoun, and
  Keutzer}]{kim2021learned}
Sehoon Kim, Sheng Shen, David Thorsley, Amir Gholami, Woosuk Kwon, Joseph
  Hassoun, and Kurt Keutzer. 2021.
\newblock \href {https://arxiv.org/abs/2107.00910} {Learned token pruning for
  transformers}.
\newblock \emph{arXiv preprint arXiv:2107.00910}.

\bibitem[{Kitaev et~al.(2020)Kitaev, Kaiser, and Levskaya}]{kitaev2019reformer}
Nikita Kitaev, Lukasz Kaiser, and Anselm Levskaya. 2020.
\newblock \href {https://openreview.net/forum?id=rkgNKkHtvB} {Reformer: The
  efficient transformer}.
\newblock In \emph{8th International Conference on Learning Representations,
  {ICLR} 2020, Addis Ababa, Ethiopia, April 26-30, 2020}. OpenReview.net.

\bibitem[{Lan et~al.(2020)Lan, Chen, Goodman, Gimpel, Sharma, and
  Soricut}]{lan2019albert}
Zhenzhong Lan, Mingda Chen, Sebastian Goodman, Kevin Gimpel, Piyush Sharma, and
  Radu Soricut. 2020.
\newblock \href {https://openreview.net/forum?id=H1eA7AEtvS} {{ALBERT:} {A}
  lite {BERT} for self-supervised learning of language representations}.
\newblock In \emph{8th International Conference on Learning Representations,
  {ICLR} 2020, Addis Ababa, Ethiopia, April 26-30, 2020}. OpenReview.net.

\bibitem[{Lang(1995)}]{lang1995newsweeder}
Ken Lang. 1995.
\newblock Newsweeder: Learning to filter netnews.
\newblock In \emph{Machine Learning Proceedings 1995}, pages 331--339.
  Elsevier.

\bibitem[{Lin et~al.(2017)Lin, Feng, dos Santos, Yu, Xiang, Zhou, and
  Bengio}]{lin2017structured}
Zhouhan Lin, Minwei Feng, C{\'{\i}}cero~Nogueira dos Santos, Mo~Yu, Bing Xiang,
  Bowen Zhou, and Yoshua Bengio. 2017.
\newblock \href {https://openreview.net/forum?id=BJC\_jUqxe} {A structured
  self-attentive sentence embedding}.
\newblock In \emph{5th International Conference on Learning Representations,
  {ICLR} 2017, Toulon, France, April 24-26, 2017, Conference Track
  Proceedings}. OpenReview.net.

\bibitem[{Liu et~al.(2020)Liu, Zhou, Wang, Zhao, Deng, and
  Ju}]{liu2020fastbert}
Weijie Liu, Peng Zhou, Zhiruo Wang, Zhe Zhao, Haotang Deng, and Qi~Ju. 2020.
\newblock \href {https://doi.org/10.18653/v1/2020.acl-main.537} {{F}ast{BERT}:
  a self-distilling {BERT} with adaptive inference time}.
\newblock In \emph{Proceedings of the 58th Annual Meeting of the Association
  for Computational Linguistics}, pages 6035--6044, Online. Association for
  Computational Linguistics.

\bibitem[{Liu et~al.(2019)Liu, Ott, Goyal, Du, Joshi, Chen, Levy, Lewis,
  Zettlemoyer, and Stoyanov}]{liu2019roberta}
Yinhan Liu, Myle Ott, Naman Goyal, Jingfei Du, Mandar Joshi, Danqi Chen, Omer
  Levy, Mike Lewis, Luke Zettlemoyer, and Veselin Stoyanov. 2019.
\newblock \href {https://arxiv.org/abs/1907.11692} {Roberta: A robustly
  optimized bert pretraining approach}.
\newblock \emph{arXiv preprint arXiv:1907.11692}.

\bibitem[{Maas et~al.(2011)Maas, Daly, Pham, Huang, Ng, and
  Potts}]{maas2011learning}
Andrew~L. Maas, Raymond~E. Daly, Peter~T. Pham, Dan Huang, Andrew~Y. Ng, and
  Christopher Potts. 2011.
\newblock \href {https://aclanthology.org/P11-1015} {Learning word vectors for
  sentiment analysis}.
\newblock In \emph{Proceedings of the 49th Annual Meeting of the Association
  for Computational Linguistics: Human Language Technologies}, pages 142--150,
  Portland, Oregon, USA. Association for Computational Linguistics.

\bibitem[{Panda et~al.(2016)Panda, Sengupta, and Roy}]{panda2016conditional}
Priyadarshini Panda, Abhronil Sengupta, and Kaushik Roy. 2016.
\newblock Conditional deep learning for energy-efficient and enhanced pattern
  recognition.
\newblock In \emph{2016 Design, Automation \& Test in Europe Conference \&
  Exhibition (DATE)}, pages 475--480. IEEE.

\bibitem[{Rumelhart et~al.(1986)Rumelhart, Hinton, and
  Williams}]{rumelhart1986learning}
David~E Rumelhart, Geoffrey~E Hinton, and Ronald~J Williams. 1986.
\newblock Learning representations by back-propagating errors.
\newblock \emph{nature}, 323(6088):533--536.

\bibitem[{Sanh et~al.(2019)Sanh, Debut, Chaumond, and
  Wolf}]{sanh2019distilbert}
Victor Sanh, Lysandre Debut, Julien Chaumond, and Thomas Wolf. 2019.
\newblock \href {https://arxiv.org/abs/1910.01108} {Distilbert, a distilled
  version of bert: smaller, faster, cheaper and lighter}.
\newblock \emph{arXiv preprint arXiv:1910.01108}.

\bibitem[{Seo et~al.(2018)Seo, Min, Farhadi, and Hajishirzi}]{seo2018neural}
Min~Joon Seo, Sewon Min, Ali Farhadi, and Hannaneh Hajishirzi. 2018.
\newblock \href {https://openreview.net/forum?id=Sy-dQG-Rb} {Neural speed
  reading via skim-rnn}.
\newblock In \emph{6th International Conference on Learning Representations,
  {ICLR} 2018, Vancouver, BC, Canada, April 30 - May 3, 2018, Conference Track
  Proceedings}. OpenReview.net.

\bibitem[{Tay et~al.(2020)Tay, Dehghani, Bahri, and Metzler}]{tay2020efficient}
Yi~Tay, Mostafa Dehghani, Dara Bahri, and Donald Metzler. 2020.
\newblock Efficient transformers: A survey.
\newblock \emph{arXiv e-prints}, pages arXiv--2009.

\bibitem[{Teerapittayanon et~al.(2016)Teerapittayanon, McDanel, and
  Kung}]{teerapittayanon2016branchynet}
Surat Teerapittayanon, Bradley McDanel, and Hsiang-Tsung Kung. 2016.
\newblock Branchynet: Fast inference via early exiting from deep neural
  networks.
\newblock In \emph{2016 23rd International Conference on Pattern Recognition
  (ICPR)}, pages 2464--2469. IEEE.

\bibitem[{Vaswani et~al.(2017)Vaswani, Shazeer, Parmar, Uszkoreit, Jones,
  Gomez, Kaiser, and Polosukhin}]{vaswani2017attention}
Ashish Vaswani, Noam Shazeer, Niki Parmar, Jakob Uszkoreit, Llion Jones,
  Aidan~N. Gomez, Lukasz Kaiser, and Illia Polosukhin. 2017.
\newblock \href
  {https://proceedings.neurips.cc/paper/2017/hash/3f5ee243547dee91fbd053c1c4a845aa-Abstract.html}
  {Attention is all you need}.
\newblock In \emph{Advances in Neural Information Processing Systems 30: Annual
  Conference on Neural Information Processing Systems 2017, December 4-9, 2017,
  Long Beach, CA, {USA}}, pages 5998--6008.

\bibitem[{Wang et~al.(2019)Wang, Singh, Michael, Hill, Levy, and
  Bowman}]{wang2018glue}
Alex Wang, Amanpreet Singh, Julian Michael, Felix Hill, Omer Levy, and
  Samuel~R. Bowman. 2019.
\newblock \href {https://openreview.net/forum?id=rJ4km2R5t7} {{GLUE:} {A}
  multi-task benchmark and analysis platform for natural language
  understanding}.
\newblock In \emph{7th International Conference on Learning Representations,
  {ICLR} 2019, New Orleans, LA, USA, May 6-9, 2019}. OpenReview.net.

\bibitem[{Wang and Zhang(2020)}]{wang2020q}
Chunpei Wang and Xiaowang Zhang. 2020.
\newblock Q-bert: A bert-based framework for computing sparql similarity in
  natural language.
\newblock In \emph{Companion Proceedings of the Web Conference 2020}, pages
  65--66.

\bibitem[{Wang et~al.(2021)Wang, Zhang, and Han}]{wang2021spatten}
Hanrui Wang, Zhekai Zhang, and Song Han. 2021.
\newblock Spatten: Efficient sparse attention architecture with cascade token
  and head pruning.
\newblock In \emph{2021 IEEE International Symposium on High-Performance
  Computer Architecture (HPCA)}, pages 97--110. IEEE.

\bibitem[{Wolf et~al.(2020)Wolf, Debut, Sanh, Chaumond, Delangue, Moi, Cistac,
  Rault, Louf, Funtowicz, Davison, Shleifer, von Platen, Ma, Jernite, Plu, Xu,
  Le~Scao, Gugger, Drame, Lhoest, and Rush}]{Wolf2019HuggingFacesTS}
Thomas Wolf, Lysandre Debut, Victor Sanh, Julien Chaumond, Clement Delangue,
  Anthony Moi, Pierric Cistac, Tim Rault, Remi Louf, Morgan Funtowicz, Joe
  Davison, Sam Shleifer, Patrick von Platen, Clara Ma, Yacine Jernite, Julien
  Plu, Canwen Xu, Teven Le~Scao, Sylvain Gugger, Mariama Drame, Quentin Lhoest,
  and Alexander Rush. 2020.
\newblock \href {https://doi.org/10.18653/v1/2020.emnlp-demos.6} {Transformers:
  State-of-the-art natural language processing}.
\newblock In \emph{Proceedings of the 2020 Conference on Empirical Methods in
  Natural Language Processing: System Demonstrations}, pages 38--45, Online.
  Association for Computational Linguistics.

\bibitem[{Wu et~al.(2020)Wu, Liu, Lin, Lin, and Han}]{wu2019lite}
Zhanghao Wu, Zhijian Liu, Ji~Lin, Yujun Lin, and Song Han. 2020.
\newblock \href {https://openreview.net/forum?id=ByeMPlHKPH} {Lite transformer
  with long-short range attention}.
\newblock In \emph{8th International Conference on Learning Representations,
  {ICLR} 2020, Addis Ababa, Ethiopia, April 26-30, 2020}. OpenReview.net.

\bibitem[{Xin et~al.(2020)Xin, Tang, Lee, Yu, and Lin}]{xin2020deebert}
Ji~Xin, Raphael Tang, Jaejun Lee, Yaoliang Yu, and Jimmy Lin. 2020.
\newblock \href {https://doi.org/10.18653/v1/2020.acl-main.204} {{D}ee{BERT}:
  Dynamic early exiting for accelerating {BERT} inference}.
\newblock In \emph{Proceedings of the 58th Annual Meeting of the Association
  for Computational Linguistics}, pages 2246--2251, Online. Association for
  Computational Linguistics.

\bibitem[{Ye et~al.(2021)Ye, Lin, Huang, and Sun}]{ye2021tr}
Deming Ye, Yankai Lin, Yufei Huang, and Maosong Sun. 2021.
\newblock \href {https://doi.org/10.18653/v1/2021.naacl-main.463} {{TR}-{BERT}:
  Dynamic token reduction for accelerating {BERT} inference}.
\newblock In \emph{Proceedings of the 2021 Conference of the North American
  Chapter of the Association for Computational Linguistics: Human Language
  Technologies}, pages 5798--5809, Online. Association for Computational
  Linguistics.

\bibitem[{Yu et~al.(2017)Yu, Lee, and Le}]{yu2017learning}
Adams~Wei Yu, Hongrae Lee, and Quoc Le. 2017.
\newblock \href {https://doi.org/10.18653/v1/P17-1172} {Learning to skim text}.
\newblock In \emph{Proceedings of the 55th Annual Meeting of the Association
  for Computational Linguistics (Volume 1: Long Papers)}, pages 1880--1890,
  Vancouver, Canada. Association for Computational Linguistics.

\bibitem[{Zaheer et~al.(2020)Zaheer, Guruganesh, Dubey, Ainslie, Alberti,
  Ontanon, Pham, Ravula, Wang, Yang et~al.}]{zaheer2020big}
Manzil Zaheer, Guru Guruganesh, Avinava Dubey, Joshua Ainslie, Chris Alberti,
  Santiago Ontanon, Philip Pham, Anirudh Ravula, Qifan Wang, Li~Yang, et~al.
  2020.
\newblock \href {https://arxiv.org/abs/2007.14062} {Big bird: Transformers for
  longer sequences}.
\newblock \emph{arXiv preprint arXiv:2007.14062}.

\bibitem[{Zhang et~al.(2015)Zhang, Zhao, and LeCun}]{zhang2015character}
Xiang Zhang, Junbo~Jake Zhao, and Yann LeCun. 2015.
\newblock \href
  {https://proceedings.neurips.cc/paper/2015/hash/250cf8b51c773f3f8dc8b4be867a9a02-Abstract.html}
  {Character-level convolutional networks for text classification}.
\newblock In \emph{Advances in Neural Information Processing Systems 28: Annual
  Conference on Neural Information Processing Systems 2015, December 7-12,
  2015, Montreal, Quebec, Canada}, pages 649--657.

\bibitem[{Zhou et~al.(2020)Zhou, Xu, Ge, McAuley, Xu, and Wei}]{zhou2020bert}
Wangchunshu Zhou, Canwen Xu, Tao Ge, Julian McAuley, Ke~Xu, and Furu Wei. 2020.
\newblock Bert loses patience: Fast and robust inference with early exit.
\newblock \emph{Advances in Neural Information Processing Systems}, 33.

\end{thebibliography}
\bibliographystyle{acl_natbib}




\end{document}